\documentclass[10pt,twocolumn,letterpaper]{article}

\usepackage[pagenumbers]{cvpr}      

\usepackage{longtable}
\usepackage{float}
\usepackage{array}
\usepackage{algorithm}
\usepackage{algorithmic}
\usepackage{tabularx}
\usepackage{multirow}
\usepackage{titletoc}
\usepackage[table]{xcolor}
\usepackage{dblfloatfix}

\definecolor{cvprblue}{rgb}{0.21,0.49,0.74}
\usepackage[pagebackref,breaklinks,colorlinks,allcolors=cvprblue]{hyperref}

\def\paperID{5} 
\def\confName{CVPR}
\def\confYear{2026}

\title{Synthetic Data Alone is Enough? \\
Rethinking Data Scarcity in Pediatric Rare Disease Recognition}

\author{Ganlin Feng$^{1}$ \quad Yuxi Long$^{1}$ \quad Erin Lou$^{2}$ \quad Lianghong Chen$^{1}$ \\ \quad Zihao Jing$^{1}$ \quad Pingzhao Hu$^{1,2,*}$ \quad Wei Xu$^{2,*}$ \vspace{0.3em} \\
{$^1$Western University} \quad
{$^2$University of Toronto} \quad 
{$^*$Correspondce: phu49@uwo.ca, wei.xu@uhn.ca} 
}

\begin{document}
\maketitle
\begin{abstract}
Children with rare genetic diseases often exhibit distinctive facial phenotypes, yet developing computer vision systems for early diagnosis remains challenging due to extreme data scarcity, privacy constraints, and limited data sharing in pediatric settings. These challenges not only hinder automated diagnosis but also restrict the availability of visual resources for clinical genetic counseling. While prior work has shown that synthetic data can augment real datasets and preserve phenotype-level semantics, it remains unclear whether synthetic data alone is sufficient for learning in ultra-low-resource pediatric settings. In this work, we study the synthetic-only regime for pediatric rare disease recognition. Under a controlled experimental setup, models are trained exclusively on phenotype-aware synthetic facial images at increasing scales. We find that synthetic-only training achieves performance comparable to real-data-only baselines at sufficient scale across multiple backbones, suggesting that high-fidelity synthetic data can approximate clinically meaningful distributions. These findings together further enable the use of synthetic pediatric facial images as privacy-preserving resources for genetic education and counseling, supporting clinician training and patient communication. Our results highlight the potential of computer vision to improve data efficiency and expand accessible visual tools in children’s healthcare.
\end{abstract}    
\section{Introduction}
\label{sec:intro}
Rare diseases affect a large population worldwide, with many conditions presenting in childhood \cite{GAHL201251}. Early diagnosis remains challenging due to heterogeneous clinical manifestations and limited clinical exposure \cite{kovac2024, baynam2016rare}. In general, rare disease diagnosis relies on multiple sources of evidence, including genetic information, clinical records (e.g., electronic health records), and dysmorphology assessment from patient appearance \cite{zurynski2017rare}. In practice, genetic information or clinical records may not always be readily available, especially in low-resource settings \cite{schieppati2008}. In contrast, patient facial images are easier to acquire and often contain visually informative patterns associated with underlying conditions, making them a practical modality for early recognition.

Recent advances in computer vision have enabled automated analysis of facial images for rare disease recognition \cite{gurovich2019, hsieh2022}. However, progress in this direction is fundamentally limited by several key bottlenecks. Pediatric rare disease datasets are limited in size, often containing only a few samples per condition. At the same time, strict privacy constraints associated with children's facial data restrict data sharing, and institutional barriers further limit dataset aggregation \cite{Kaissis2020SecurePA}. These challenges create a highly constrained data regime. 

As a result, these limitations not only hinder the development of deep learning models, but also restrict access to visual examples for human learning and genetic education. Clinicians and trainees therefore have limited exposure to diverse phenotypic variations, making it difficult to recognize rare conditions in practice, particularly under strict privacy constraints.

A promising direction to address these challenges is the use of synthetic data \cite{doi:10.1148/radiol.232471}. Prior work has shown that synthetic facial images can preserve phenotype-level semantic information \cite{kirchhoff2025}. However, realism alone does not guarantee usability. It remains unclear whether using synthetic data only is sufficient to support machine learning, particularly in ultra-low-resource pediatric settings.

In this work, we address this question by studying the synthetic-only regime for pediatric rare disease recognition. Specifically, we study whether models trained exclusively on synthetic facial images can effectively learn disease-relevant visual patterns in the absence of real data. To this end, we conduct a systematic evaluation across multiple backbone architectures and varying synthetic data scales to highlight both the potential and limitations of synthetic data for learning in ultra-low-resource settings.

Overall, our contributions are summarized as follows:
\begin{itemize}
    \item We provide a systematic study of the \textbf{synthetic-only regime} for pediatric rare disease recognition, isolating the role of synthetic data without mixing real samples.
    \item We analyze the behavior of \textbf{synthetic data scaling}, evaluating performance across multiple data sizes and backbone architectures.
    \item We discuss the implications of synthetic data as a \textbf{privacy-preserving visual resource}, enabling future applications such as genetic education and clinical training.
\end{itemize}
\section{Related Work}
\label{sec:related_work}

\noindent\textbf{Facial Analysis for Rare Disease Recognition.}
Recent advances in computer vision have enabled automated analysis of facial images for rare disease recognition \cite{hsieh2022,gurovich2019}. Prior work has leveraged deep learning models to identify syndromic patterns from patient photos, demonstrating the potential of image-based approaches for assisting clinical diagnosis \cite{jin2020, sherif2024}. However, most existing methods rely on relatively well-represented conditions with sufficient training data, and their performance degrades significantly in ultra-low-resource settings where only a few samples are available per disease. In addition, the availability of pediatric facial datasets is severely constrained due to privacy concerns and limited data sharing, which further restricts the scalability of such approaches.

\noindent\textbf{Synthetic Data in Medical Imaging.}
To address data scarcity, synthetic data generation has been widely explored in medical imaging, particularly with generative adversarial networks and diffusion-based models \cite{wolleb2022diffusion, KAZEMINIA2020101938}. These approaches aim to improve model performance through data augmentation or to enhance data diversity. Prior studies have shown that synthetic images can achieve high visual fidelity and preserve semantic information relevant to downstream tasks \cite{feng2026rdface,kirchhoff2025}. However, existing work primarily focuses on using synthetic data as a supplement to real data. It remains unclear whether synthetic data alone is sufficient to support effective learning, especially in low-data regimes.

\noindent\textbf{Visual Resources for Learning and Clinical Training}
In clinical practice, visual examples play an important role in learning to recognize rare disease phenotypes \cite{koretzky2016, ren2021}. However, the availability of shareable patient images is limited due to privacy constraints, particularly for pediatric populations \cite{Price2019PrivacyIT}. While digital tools and datasets have been developed to support medical education, there remains a lack of scalable and privacy-preserving visual resources.
\section{Methodology}
\label{sec:methodology}

\subsection{Dataset and Data Split}
We use RDFace-Syn, a synthetic dataset proposed by the RDFace benchmark dataset \cite{feng2026rdface}, where images are generated using DreamBooth \cite{ruiz2023}, a diffusion-based generation approach, as described in their original work. The synthetic data is generated on a per-class basis from limited real samples from RDFace, providing a scalable training set under extreme low-data conditions. Overall, RDFace-Syn contains 10,300 synthetic images across 103 rare genetic conditions, with
100 synthetic images per condition. We adopt the same test set as the original benchmark, ensuring a consistent and fair evaluation protocol. 

\subsection{Synthetic Data Setup}
We focus on the \textbf{synthetic-only} training regime, where models are trained exclusively on generated data without access to real images during training. This setup allows us to isolate the effect of synthetic data without relying on real training samples or modifying the generation process. Following the RDFace-Syn protocol, synthetic samples within each class are ranked based on their cosine similarity to real class prototypes, computed using facial landmark representations. We construct training sets of increasing scale by selecting the top-$n$ ranked samples per class, allowing us to systematically study scaling behavior with respect to synthetic data quality and quantity.

\subsection{Training Protocol}
We adopt a unified training pipeline across all experiments. Multiple backbone architectures are trained under identical settings, differing only in the size of the synthetic training set. Input images are resized to $224 \times 224$ and normalized using ImageNet \cite{deng2009} statistics. For each backbone, the final classification layer is replaced with a 103-way softmax corresponding to disease classes. Additionally, 5-fold cross-validation is used for hyperparameter selection on the training data. For real-data-only baselines, the same training protocol is followed, using the real training set from RDFace.

\subsection{Evaluation}
Model performance is evaluated on the held-out real-image test set using Top-$k$ classification accuracy. A prediction is considered correct if the ground-truth label appears within the top-$k$ predicted classes. Results are reported as mean and standard deviation across cross-validation folds.
\renewcommand{\arraystretch}{0.9} 
\setlength{\tabcolsep}{8pt}
\begin{table*}[hbtp]
\centering
\caption{Top-$k$ accuracies (\%) across backbones and synthetic cutoffs of DreamBooth-only samples. The raw in \colorbox{gray!25}{gray} is the results using real data only in each setting. All results are reported as mean (standard deviation). The best synthetic results are highlighted in \textbf{bold}.}
\label{tab:topk_aug_dreambooth}
\small
\begin{tabularx}{\textwidth}{l*{7}{>{\centering\arraybackslash}X}}

\toprule
\textbf{ACC (\%)} & \textbf{Top-$n$} & \textbf{ResNet} & \textbf{DenseNet} & \textbf{FaceNet} & \textbf{VGG} & \textbf{Swin-T} & \textbf{CLIP} \\
\midrule

\rowcolor{gray!25}
\multirow{7}{*}{Top-1}
    & \textbf{Real-only}         &   6.90 (1.45)     &    15.93 (2.34)    &    9.91 (1.81)     &   11.68 (1.58)    &    14.34 (2.61)    &    3.10 (1.48)     \\
    & Top-2000   &   3.89 (1.34)    &    7.79 (1.92)    &    8.14 (2.37)    &   7.61 (1.84)    &    6.55 (0.79)    &    6.19 (1.77)    \\
    & Top-4000   &   5.32 (1.66)    &    8.85 (2.08)    &    12.74 (2.31)    &   9.20 (1.48)    &    8.14 (1.15)   &    9.38 (1.61)   \\
    & Top-6000   &   5.66 (1.01)    &    9.38 (1.01)    &    12.04 (1.61)    &   \textbf{10.62 (1.40)}  &    9.03 (1.31)   &    \textbf{10.62 (2.26)}   \\
    & Top-8000  &  \textbf{7.08 (1.24)}   &    \textbf{11.50 (1.07)}   &   \textbf{13.27 (2.23)}   &   9.73 (1.08)    &    \textbf{12.39 (1.12)}   &    7.96 (2.37)   \\ 
    & Top-10000  &  3.54 (1.01)   &    4.42 (2.61)   &   9.73 (1.01)   &   7.08 (1.61)    &    5.31 (1.81)   &    5.31 (1.07) \\

\midrule

\rowcolor{gray!25}
\multirow{7}{*}{Top-5}
    & \textbf{Real-only}         &   18.58 (3.00)    &    33.63 (3.70)    &    24.60 (5.43)    &    29.91 (2.68)    &    26.19 (2.68)    &   12.74 (2.84)    \\
    & Top-2000   &   15.22 (4.98)    &    21.77 (2.70)    &    20.53 (3.72)    &    24.42 (1.73)    &    17.88 (1.31)    &   17.17 (1.34)    \\
    & Top-4000   &   15.40 (1.48)    &    22.65 (1.61)    &    20.53 (3.72)    &    25.49 (2.46)    &    20.88 (1.01)    &   19.12 (2.04)    \\
    & Top-6000   &   16.64 (2.37)    &     24.42 (1.01)   &    24.60 (1.31)    &    25.31 (2.04)    &    \textbf{22.83 (1.31)}    &   21.42 (3.56)    \\
    & Top-8000  &  \textbf{20.35 (2.04)}   &    \textbf{28.32 (1.01)}   &   \textbf{29.20 (1.21)}   &    \textbf{27.43 (2.87)}    &    22.12 (2.61)   &    \textbf{23.01 (1.70)}   \\ 
    & Top-10000  &  12.39 (2.34)   &    15.93 (1.66)   &   18.58 (1.19)   &   19.47 (0.79)    &    16.81 (1.48)   &    12.39 (1.01) \\
\midrule

\rowcolor{gray!25}
\multirow{7}{*}{Top-10}
    & \textbf{Real-only}         &   28.50 (3.67)    &    43.01 (2.63)    &    34.87 (4.75)    &    38.41 (1.34)    &    35.93 (3.17)    &   19.12 (5.18)    \\
    & Top-2000   &   23.36 (4.54)    &    29.20 (0.63)    &    29.91 (4.57)    &    36.11 (2.76)    &    30.09 (2.87)    &   26.19 (2.22)    \\
    & Top-4000   &   24.07 (2.61)    &    30.62 (4.13)    &    29.91 (4.57)    &    36.11 (1.92)    &    28.32 (1.66)    &   26.37 (2.20)    \\
    & Top-6000   &   25.13 (1.01)    &   31.68 (1.70)     &    34.16 (1.01)    &    37.52 (2.04)    &    30.80 (2.29)    &   27.61 (3.45)    \\
    & Top-8000  &  \textbf{29.20 (2.55)}   &    \textbf{34.51 (1.92)}   &   \textbf{37.17 (1.34)}   &   \textbf{38.05 (1.80)}    &    \textbf{32.74 (0.63)}   &    \textbf{28.32 (2.46)}   \\ 
    & Top-10000  &  17.70 (1.07)   &    23.89 (3.17)   &   27.43 (2.76)   &   24.78 (1.66)    &    25.66 (1.92)   &    18.58 (1.92) \\
\midrule

\rowcolor{gray!25}
\multirow{7}{*}{Top-30}
    & \textbf{Real-only}         &   54.34 (2.39)    &    64.42 (1.92)    &    58.23 (5.06)    &    60.88 (2.02)    &    58.41 (3.81)    &   42.30 (4.40)    \\
    & Top-2000   &   47.43 (2.55)    &   54.69 (2.11)    &    55.58 (4.82)    &    66.73 (3.99)    &    53.27 (1.70)    &   45.31 (2.89)    \\
    & Top-4000   &   47.96 (3.78)    &   51.86 (1.73)     &    55.75 (5.77)    &    64.42 (2.37)    &    55.58 (0.97)    &   48.50 (3.93)    \\
    & Top-6000   &   \textbf{50.62 (2.02)}    &   55.93 (2.46)    &    \textbf{57.88 (4.71)}   &    \textbf{66.37 (2.58)}    &    60.88 (5.43)    &   47.96 (1.70)    \\
    & Top-8000  &  50.44 (1.65)   &    \textbf{56.64 (1.92)}   &   54.87 (1.07)   &   61.06 (1.08)    &    \textbf{62.12 (3.15)}   &    \textbf{53.98 (1.31)}   \\ 
    & Top-10000  &  36.28 (2.02)   &    43.36 (2.08)   &   45.13 (1.48)   &   46.90 (2.29)    &    46.02 (2.11)   &    46.90 (2.87) \\
\bottomrule
\end{tabularx}
\end{table*}

\section{Experiments}
\label{sec:experiments}
\subsection{Experimental Setup}
We evaluate the effectiveness of synthetic data under the synthetic-only regime across multiple backbone architectures, including ResNet152 \cite{he2016}, DenseNet169 \cite{huang2017}, FaceNet \cite{schroff2015}, VGG16 \cite{simonyan2015}, Swin-T \cite{liu2021}, and CLIP (ViT-B/32) \cite{radford2021}, under a unified experimental protocol. We consider two training regimes: (1) synthetic-only training, where models are trained exclusively on synthetic data at varying scales, and (2) real-data-only baselines, where models are trained using the real images from RDFace benchmark. To analyze scaling behavior, we construct synthetic training sets of increasing size (2K, 4K, 6K, 8K, and 10K samples) using the top-ranked synthetic images. All models are evaluated on the same real-image test set, ensuring a controlled comparison across different settings. For evaluation metrics, we report Top-1, Top-5, Top-10, and Top-30 accuracy.

\begin{figure}[h]
    \centering
    \includegraphics[width=0.9\linewidth]{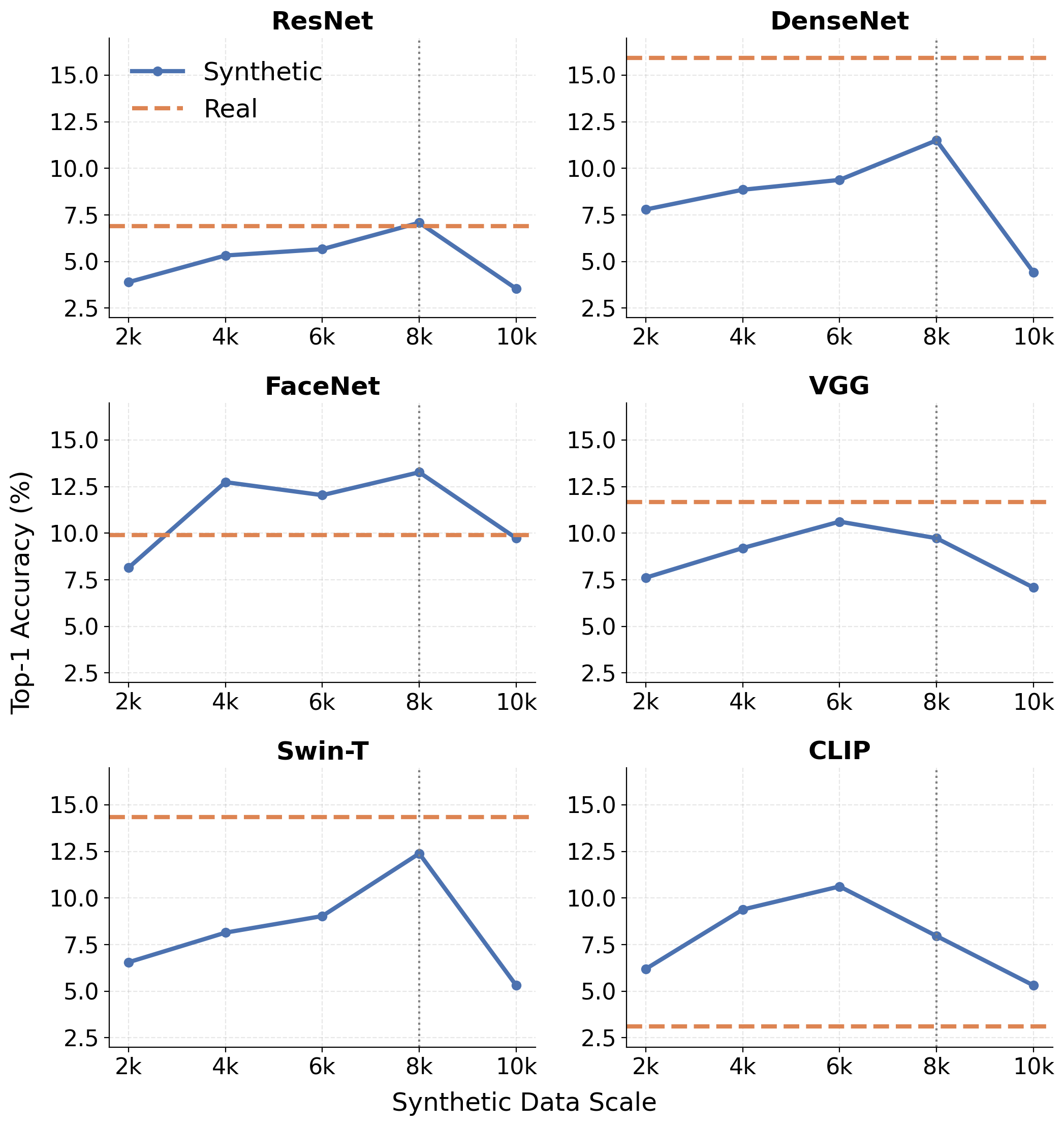}
    \caption{\textbf{Top-1 accuracy versus synthetic data scale for six backbones}. Dashed orange lines indicate real-only baselines. Performance improves with scale and varies across backbones.}
    \label{fig:trend}
\end{figure}

\begin{figure*}[t]
    \centering
    \includegraphics[width=0.9\linewidth]{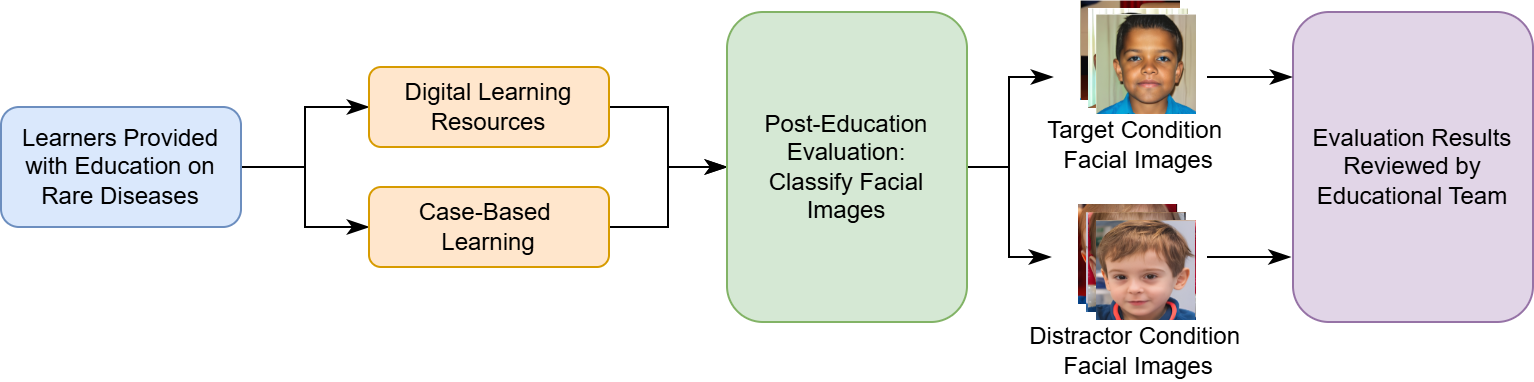}
    \caption{\textbf{Educational workflow for rare disease}. Learners are provided with resources and tasked with classifying facial images.}
    \label{fig:counseling}
\end{figure*}

\subsection{Synthetic-Only vs Real-Only Performance}
We first compare synthetic-only training against real-data-only baselines. As shown in \Cref{tab:topk_aug_dreambooth}, synthetic-only models achieve performance comparable to real-data-only baselines under several backbone architectures. We next analyze model performance under Top-1 accuracy across varying Top-$n$ settings.

Under Top-1 accuracy, synthetic-only training matches or exceeds real-only performance for certain models (\Cref{fig:trend}). For example, FaceNet and CLIP surpass their real-data baselines at larger scales (FaceNet: 13.27\% vs. 9.91\%; CLIP: 10.62\% vs. 3.10\%), while Swin-T and VGG achieve closely comparable performance to their real-only counterparts. In contrast, other architectures such as DenseNet still exhibit noticeable performance gaps (11.50\% vs. 15.93\%). Similar patterns are observed across higher Top-$k$ accuracies. Under Top-5, 10, 30 accuracy, synthetic-only performance consistently approaches real-data baselines and, in some cases, achieves competitive results depending on the backbone. Overall, these results indicate that synthetic data can serve as a viable alternative to real data under appropriate conditions, while its effectiveness remains dependent on the choice of model architecture. 

\subsection{Scaling Behavior of Synthetic Data}
We further analyze how performance varies with the amount of synthetic data. Model performance improves consistently as the number of synthetic samples increases from 2K to intermediate scales across all backbone architectures. Across all backbone architectures, performance typically peaks around 6K–8K samples. As shown in \Cref{fig:trend}, Top-1 accuracy reaches its maximum at 8K for most models (e.g., ResNet: 7.08\%,DenseNet: 11.50\%, FaceNet: 13.27\%, Swin-T: 12.39\%). 
Beyond this range, performance often decreases at larger scales (e.g., 10K), suggesting that excessive synthetic data may introduce redundancy or lower-quality samples that negatively impact learning. This behavior may be attributed to the cosine similarity-based ranking strategy, where lower-ranked samples are less representative and can degrade the training signal. Overall, these results indicate a clear scaling pattern, where increasing synthetic data is beneficial up to a moderate range but becomes less effective beyond it.
\section{Genetic Counseling and Education}
\label{sec:counseling}

Many rare pediatric genetic disorders manifest as distinctive craniofacial phenotypes \cite{kovac2024}, where observable facial features reflect underlying genetic variation. Such phenotype–genotype associations are well established in clinical genetics, providing a biological basis for using facial features as visual indicators of genetic conditions \cite{10.3389/fgene.2018.00462, doi:10.1073/pnas.1708207114}. Based on prior work showing synthetic images preserve specific phenotypes \cite{kirchhoff2025} and our findings that synthetic data can approximate real-data performance, synthetic images can serve as a practical resource for genetic counseling.

Building on this, RDFace-Syn enables targeted applications in pediatric clinical genetics education and rare disease awareness. By providing condition-specific synthetic facial images, it can be integrated into digital learning platforms and case-based curricula to support early diagnostic training in children. We provide a potential workflow (\Cref{fig:counseling}) in which educators design exercises for trainees to classify facial phenotypes, compare classification with annotations, and track diagnostic accuracy, reinforcing phenotype–disease associations and improving recognition skills and confidence in clinicians \cite{waikel2024recognition}.

In genetic counseling, synthetic images can serve as visual references to support communication with patients and families, enhancing understanding of disease-specific craniofacial features in pediatric populations \cite{https://doi.org/10.1002/jgc4.1862}. These synthetic images could be incorporated into patient-facing materials to complement general awareness tools, providing condition-specific visuals that support lay understanding.

Importantly, the use of synthetic facial images mitigates patient privacy concerns by reducing reliance on identifiable real-world data. This supports broader sharing and reuse of visual materials for training, evaluation, and public-facing education, which also aligns with ongoing research exploring synthetic data generation to enhance clinical training while ensuring ethical data sharing \cite{PEZOULAS20242892}.
\section{Conclusion}
\label{sec:conclusion}
In this work, we systematically investigate the synthetic-only regime for pediatric rare disease recognition, evaluating whether models trained exclusively on synthetic facial images can approximate real-data performance. Our findings show that synthetic-only training can achieve competitive performance under appropriate conditions, with a clear scaling behavior where performance improves with increasing data and peaks at intermediate scales. 

At the same time, we identify several limitations. Performance remains strongly dependent on model architecture, and synthetic data does not yet fully replace real data across all settings. In addition, performance degradation at larger scales suggests redundancy or noise in current generation and selection. Beyond model performance, our findings and prior work suggest that synthetic facial data can serve as meaningful representations of rare diseases, enabling genetic education and clinical training. Future work should focus on improving the fidelity of synthetic data, particularly at larger scales, and developing architecture-aware training strategies that align synthetic data with different backbone models. Moreover, clinical user study validation is needed to further assess the impact of synthetic data in genetic education and counseling workflows.
{
    \small
    \bibliographystyle{ieeenat_fullname}
    \bibliography{main}
}


\end{document}